\documentclass{article}

\PassOptionsToPackage{numbers, compress}{natbib}
\usepackage[preprint]{neurips_2024}

\usepackage[utf8]{inputenc} 
\usepackage[T1]{fontenc}    
\usepackage[hidelinks]{hyperref}       
\usepackage{url}            
\usepackage{booktabs}       
\usepackage{amsfonts}       
\usepackage{nicefrac}       
\usepackage{microtype}      
\usepackage{xcolor}         

\usepackage{times}
\usepackage{soul}
\usepackage{url}
\usepackage[small]{caption}
\usepackage{subcaption} 
\usepackage{graphicx}
\usepackage{microtype}
\usepackage{subfloat}
\usepackage{amsmath}        
\usepackage{amssymb}
\usepackage{mathtools}
\usepackage{amsthm}
\usepackage[nohyperlinks]{acronym}
\usepackage{xspace}         
\usepackage{enumitem}       
\usepackage{multirow}       
\usepackage[textsize=tiny]{todonotes}
\makeatletter
\DeclareRobustCommand\onedot{\futurelet\@let@token\@onedot}
\def\@onedot{\ifx\@let@token.\else.\null\fi\xspace}

\def\eg{\emph{e.g}\onedot} 
\def\ie{\emph{i.e}\onedot}

\def\etal{\emph{et al}\onedot}
\makeatother

\DeclareMathOperator*{\argmax}{arg \ max}

\newcommand*{\method}{UniTable}

\definecolor{structure}{RGB}{0, 133, 255} 
\definecolor{bbox}{RGB}{227, 164, 0} 
\definecolor{content}{RGB}{19, 139, 0} 
\acrodef{tsr}[TR]{table recognition}
\acrodef{teds}[TEDS]{tree-edit-distance-based similarity}
\acrodef{vlp}[VLP]{vision-language pretraining}
\acrodef{vlm}[VLM]{vision-language models}
\acrodef{cnn}[CNN]{convolutional neural network}
\acrodef{rnn}[RNN]{recurrent neural network}
\acrodef{pp}[pp]{percentage points}
\acrodef{sota}[SOTA]{state-of-the-art}
\acrodef{vit}[ViT]{vision transformer}
\acrodef{relu}[ReLU]{rectified linear unit}
\acrodef{gelu}[GELU]{gaussian error linear unit}
\acrodef{nlp}[NLP]{natural language processing}
\acrodef{vilt}[ViLT]{vision-and-language transformer}
\acrodef{ssl}[SSL]{self-supervised learning}
\acrodef{rf}[RF]{receptive field}
\acrodef{iou}[IoU]{intersection over union}
\acrodef{mac}[MAC]{Multiply-Add Operations per Second}
\acrodef{edd}[EDD]{encoder-dual-decoder}
\acrodef{pdf}[PDF]{portable document format}
\acrodef{ssp}[SSP]{self-supervised pretraining}
\acrodef{mim}[MIM]{masked image modeling}
\acrodef{mlm}[MLM]{masked language modeling}
\acrodef{vqvae}[VQ-VAE]{Vector Quantized-Variational AutoEncoder}
\acrodef{ocr}[OCR]{optical character recognition}
\acrodef{bbox}[bbox]{bounding box}
\acrodef{ap}[AP]{average prevision}
\acrodef{map}[mAP]{mean \acs{ap}}
\acrodef{icdar19}[IC19B2M]{ICDAR 2019 B2 Modern}
\acrodef{car}[CAR]{cell adjacency relations}
\acrodef{wf1}[WAvg. F1]{weighted average F1}
\acrodef{agi}[AGI]{artificial general intelligence}
\title{UniTable: Towards a Unified Framework for Table  Recognition via Self-Supervised Pretraining}

\author{%
  ShengYun Peng$^{1}$ \quad Aishwarya Chakravarthy$^{1}$ \quad Seongmin Lee$^{1}$ \quad
  Xiaojing Wang$^{2}$\\
  \textbf{Rajarajeswari Balasubramaniyan}$^{2}$ \quad \textbf{Duen Horng Chau}$^{1}$\\
  $^1$Georgia Tech \quad $^2$ADP, Inc.\\
  \texttt{\{speng65,achakrav6,seongmin\}@gatech.edu}\\
  \texttt{\{xiaojing.wang,raji.balasubramaniyan\}@adp.com}
}

\begin{document}
\maketitle

\begin{abstract}
Tables convey factual and quantitative data with implicit conventions created by humans that are often challenging for machines to parse. 
Prior work on \ac{tsr} has mainly centered around complex task-specific combinations of available inputs and tools. 
We present \textbf{\method{}},
a training framework that unifies both the \textbf{training paradigm} and \textbf{training objective} of \ac{tsr}.
Its training paradigm combines the simplicity of purely pixel-level inputs with the effectiveness and scalability empowered by \ac{ssp} from diverse unannotated tabular images.
Our framework unifies the training objectives of all three \ac{tsr} tasks --- extracting table structure, cell content, and cell \ac{bbox} --- into a unified task-agnostic training objective: language modeling. 
Extensive quantitative and qualitative analyses highlight \method{}'s \ac{sota} performance on four of the largest \ac{tsr} datasets.
\method{}'s table parsing capability has surpassed both existing \ac{tsr} methods and general large \acp{vlm}, \eg, GPT-4o, GPT-4-turbo with vision, and LLaVA. 
Our code is publicly available at {\footnotesize 
\url{https://github.com/poloclub/unitable}}, featuring a Jupyter Notebook that includes the complete inference pipeline, fine-tuned across multiple \ac{tsr} datasets, supporting all three \ac{tsr} tasks.
\end{abstract}
\section{Introduction}
\label{sec:intro}

Tables are ubiquitous in documents, as they serve to summarize factual and quantitative data --- information that is cumbersome to describe in text but nevertheless crucial~\cite{gobel2013icdar, gao2019icdar}. 
Due to the implicit conventions used by humans in creating tables, the representations within tables are often challenging for machines to parse. 
Even the milestone \ac{vlm}, GPT-4o~\cite{gpt4o}, GPT-4V(vision)~\cite{yang2023dawn} and LLaVA~\cite{liu2024visual}, still struggles with various document-related tasks including \ac{tsr}. 
GPT-4V tends to omit content in large tables and performs worse when faced with complex tables, \eg, spanning or empty cells and uneven text distributions~\cite{shi2023exploring}. 

Prior work on \ac{tsr} has mainly centered around complex task-specific combinations of available inputs and tools. 
Typically, table structure was predicted by an image-to-text pipeline~\cite{zhong2020image} and cell \ac{bbox} was predicted by a detection head, \eg, Faster R-CNN~\cite{smock2022pubtables} or DETR~\cite{nassar2022tableformer}. 
The assumption regarding predicting cell content varies: some studies assume the presence of a \ac{pdf} accompanying the tabular image~\cite{nassar2022tableformer}, while others rely on external text line detection and text recognition models~\cite{ye2021pingan, huang2023improving}. 
However, training generic Transformers under a single language modeling objective, \ie, predicting the next token, for diverse tasks has achieved remarkable success across diverse tasks in language~\cite{radford2019language, brown2020language}, vision~\cite{chen2021pix2seq, chen2022unified}, and vision-and-language domains~\cite{achiam2023gpt}. 
We wonder whether understanding visual language in tables, \ie, extracting table structure, cell content, and cell \ac{bbox} from tabular images, can be seamlessly integrated into the language modeling training framework.
This integration is challenging because: 
(1) the fusion of vision and language in tabular images demands high-fidelity reading and rich high-level representations~\cite{lee2023pix2struct}; 
(2) the large amount of unannotated tabular images in practice cannot be leveraged by existing supervised learning approaches~\cite{nassar2022tableformer, ma2023robust}; 
(3) the diverse output of table-related tasks are typically addressed by task-specific models~\cite{huang2023improving}; and
(4) direct application of a linear projection Transformer results in the significant performance drop~\cite{peng2023high}, leading prior work to exclusively employ a hybrid \ac{cnn}-Transformer architecture. 
We resolve the above challenges by proposing \textbf{\method{}}, a training framework that unifies both \textbf{training paradigm} and \textbf{training objective} of \ac{tsr} and make the following major contributions (Fig.~\ref{fig:pipeline}):

\begin{figure}[t]
\centering
\includegraphics[width=0.93\linewidth]{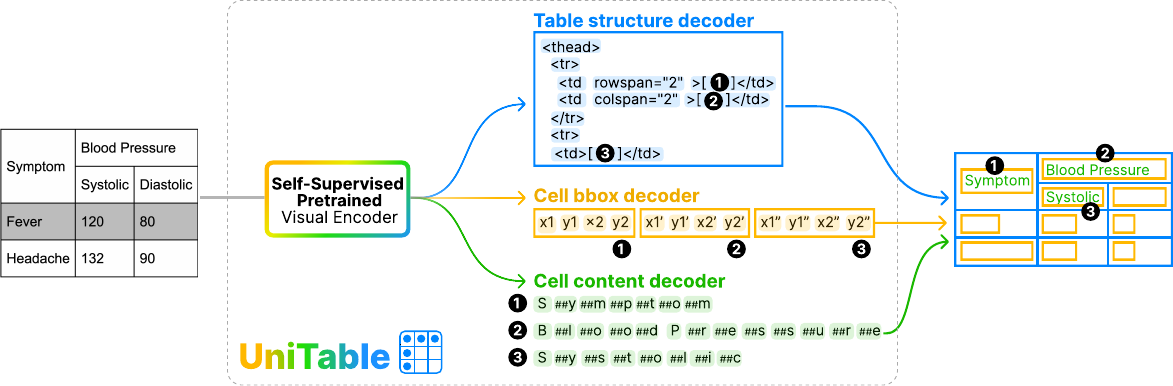}
\caption{\textbf{\method{}}, a training framework that unifies both \textbf{training paradigm} and \textbf{training objective} of \acs{tsr}. 
In \method{}, the visual encoder is self-supervised pretrained and then finetuned along with the task decoder on supervised datasets.
\method{} unifies the training objectives of all three \ac{tsr} tasks --- extracting \textcolor{structure}{table structure}, \textcolor{bbox}{cell \acs{bbox}}, and \textcolor{content}{cell content} --- into a unified task-agnostic training objective: language modeling.
With \method{}, the user inputs a tabular image and obtains the corresponding digitalized table in HTML. 
}
\label{fig:pipeline}
\end{figure}

\begin{figure}[b]
\centering
\includegraphics[width=0.93\linewidth]{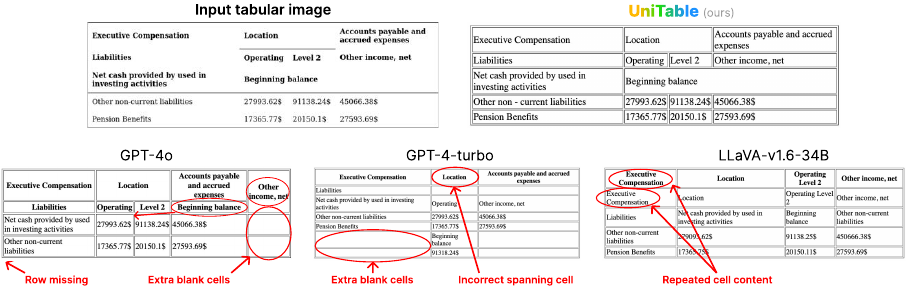}
\caption{The table parsing capability of \method{} has surpassed that of general large \acp{vlm}, \eg, GPT-4o, GPT-4-turbo with vision, and LLaVA. 
For complex tables that include multiple spanning cells, \method{} can successfully reconstruct the table in HTML, whereas general large \acp{vlm} fail in various aspects. 
}
\label{fig:vlm}
\end{figure}

\begin{enumerate}[leftmargin=*,topsep=0pt]
\itemsep0em 
\item \textbf{\method{}'s training paradigm combines the simplicity of purely pixel-level inputs with the effectiveness and scalability empowered by \ac{ssp} from diverse unannotated tabular images.}
Specifically, \method{} unifies the training paradigm for \ac{tsr}: pretraining the visual encoder by predicting the masked tabular images in a self-supervised manner and finetuning the visual encoder along with the task decoder on supervised datasets.
With \method{}, the table structure prediction on
SynthTabNet~\cite{nassar2022tableformer}, a comprehensive dataset with 600k tables across finance, marketing, and academia in both dense and sparse format, 
achieves the \ac{sota} 99.18\% when self-supervised pretrained on 2M images, 
significantly lifting the original accuracy of 84.04\% when trained from scratch using a linear projection Transformer.
Owing to the powerful \ac{ssp}, \method{} has also successfully mitigated the performance drop caused by replacing the \ac{cnn} backbone with the linear projection. 

\item \textbf{\method{} unifies the training objectives of all three \ac{tsr} tasks --- extracting table structure, cell content, and cell \ac{bbox} --- into a unified task-agnostic training objective: language modeling.}
Specifically, the input to our model is an image in the form of raw pixels only and the output is text in the form of token sequences, and the training objective is language modeling. 
\method{}'s \ac{sota} performance on the FinTabNet dataset~\cite{zheng2021global} demonstrates our approach's generalizability to the \ac{pdf} input modality as we can simply convert \ac{pdf} to images. 
Our framework also enables us to leverage the power of \ac{ssp} on large-scale unannotated tabular images as all models are finetuned from \ac{ssp}.
\method{}'s
unified training objective applies to both linear projection Transformer and hybrid CNN-Transformer architectures conventionally used in TSR. 

\item \textbf{Extensive quantitative and qualitative analyses highlight \method{}'s \ac{sota} performance on four of the largest \ac{tsr} datasets}:
 ICDAR 2019 B2 Modern~\cite{gao2019icdar}, PubTabNet~\cite{zhong2020image}, FinTabNet~\cite{zheng2021global}, and SynthTabNet~\cite{nassar2022tableformer}. 
\method{}'s table parsing capability has surpassed both existing \ac{tsr} methods and general large \acp{vlm} (Fig.~\ref{fig:vlm}), \eg, GPT-4o, GPT-4V(ision), and LLaVA.
Due to our unified language modeling framework formulation, we discover three types of previously unacknowledged inconsistencies in the groundtruth annotations of PubTables-1M ~\cite{smock2022pubtables}, one of the largest TSR datasets, 
accounting for more than 53.10\% of its training set.
Our visualization of the visual tokens reveals the key reason for why \ac{ssp} works --- the visual semantics captured by the visual codebook show a fine-grained categorization to represent the implicit human conventions used when creating the tables.  

\item \textbf{Open-source code and \method{} in practice.}
To promote reproducible research, enhance transparency, \ac{sota} innovations, and facilitate fair comparisons in our domain as tables are a promising modality for representation learning,
we open-source our code (anonymized)\footnote{\scriptsize 
\url{https://github.com/poloclub/unitable}
}.
We provide all the details regarding training, validation, testing, and ablation studies. 
To enable users to easily try \method{} on their own tabular images and obtain fully digitized HTML tables,
we release the first-of-its-kind Jupyter Notebook of the whole inference pipeline, fine-tuned across multiple \ac{tsr} datasets, supporting all three \ac{tsr} tasks.
%

\end{enumerate}

\section{Background}
\label{sec: background}

\subsection{Task Definition}
\label{sec: definition}
The goal of \ac{tsr} is to translate the input tabular image $\mathbf{I}$ into a machine-readable sequence $\mathbf{T}$, which typically consists of table structure $\mathbf{S}$, table cell \ac{bbox} $\mathbf{B}$, and table cell content $\mathbf{C}$. 
Structure $\mathbf{S} = [s_1, ..., s_m]$ is a sequence of tokenized HTML table tags $s$, cell \ac{bbox} $\mathbf{B} = [\mathbf{b}_1, ..., \mathbf{b}_n]$ is a sequence of bboxes defined by $\mathbf{b} = (x_{min}, y_{min}, x_{max}, y_{max})$, and cell content $\mathbf{C} = [c_1, ..., c_n]$ comprises the content within each cell in reading order. 
Note that the sequence length of $\mathbf{B}$ and $\mathbf{C}$ are the same, but shorter than $\mathbf{S}$ as the HTML tags contain both empty and non-empty cells. 
Since each cell $i$ is defined by a single \ac{bbox}, $c_i$ can have either single or multiple text lines. 

\subsection{Model Architecture}
\label{sec: arch}
\Ac{tsr} model architecture has two modules: visual encoder and task decoder. 
The visual encoder extracts features from input $\mathbf{I}$, and the task decoder predicts $\mathbf{T}$. 
For the visual encoder, prior work employed either an off-the-shelf \ac{cnn} backbone, \eg, ResNets and ResNet variants~\cite{he2016deep}, or hybrid \ac{cnn}-Transformer architecture. 
EDD~\cite{zhong2020image} explored five different ResNet-18 variants, TableMaster~\cite{lu2021master} combined residual blocks with multi-aspect global context attention, TableFormer~\cite{nassar2022tableformer} connected a ResNet-18 with Transformer encoder layers, and VAST~\cite{huang2023improving} adopted the first four stages of a ResNet-31. 
These convolution layers cannot be replaced because a direct employment of a vanilla Transformer with linear projection leads to a significant performance drop~\cite{peng2023high}. 
However, linear projection, that divides image into patches, is a widely used input image processor in \ac{sota} vision Transformers~\cite{dosovitskiy2020image, liu2021swin}, \acp{vlm}~\cite{li2022blip, liu2023visual} and multi-modal models~\cite{mizrahi20234m}.
Recent work on visual language understanding has also successfully adopted the linear projection~\cite{kim2022ocr, lee2023pix2struct}. 
Thus, to avoid having a separate architecture design solely for the table domain, we aim to keep the linear projection Transformer, and mitigate the performance gap by \ac{ssp}. 
Sec.~\ref{sec: method} shows that pretraining the visual encoder in a self-supervised manner significantly helps the model learn how to parse the table structure and achieves performance even higher than architectures with convolutions.


\section{A Unified Framework for Pretraining and Finetuning TR Models}
\label{sec: method}

We pretrain the visual encoder by predicting the masked tabular images in a self-supervised manner and finetune the visual encoder along with the task decoder using the supervised dataset for each task, as shown in Fig.~\ref{fig:pipeline}. 
Sec.~\ref{sec: method-ssp} introduces the \ac{ssp} of the visual encoder, and Sec.~\ref{sec: method-unify} details how we unify all \ac{tsr} tasks and finetune with the pretrained visual encoder. 

\subsection{Self-Supervised Pretraining of the Visual Encoder}
\label{sec: method-ssp}

Before pretraining, each tabular image $I$ is tokenized into discrete visual tokens. 
During pretraining, $I$ is divided into patches, and a portion of the image patches are masked so that the visual encoder predicts which visual token is chosen to replace the masked regions. 

\textbf{Image tokens.}
Define a visual codebook in the latent space $Z \in \mathbb{R}^{K \times D}$, representing $K$ entries of visual tokens $z_i \in \mathbb{R}^D$. 
The visual codebook is trained in such a way that an input image, once it is embedded into an image grid, each pixel on the embedded image grid can be substituted with a visual token from the codebook. 
Decoding this modified image grid will then reconstruct the input image.
We use \ac{vqvae}~\cite{van2017neural} to train the visual codebook. 
Specifically, the tabular image $I$ is tokenized into discrete tokens $z$ after passing through the encoder $q_{\phi}(z | I)$, and the decoder $p_{\psi}(I | z)$ takes these discrete tokens and rebuilds the original image. 
The training objective of \ac{vqvae} is to maximize the $\mathbb{E}_{z \sim q_{\phi}(z | I)} [ \log p_{\psi}(I | z) ]$ with respect to $\phi$ and $\psi$.
The training is non-differentialble due to the categorical distribution in selecting visual tokens. 
Thus, we use Gumbel-Softmax~\cite{jang2016categorical} as a reparameterization trick following DALL-E~\cite{ramesh2021zero}.
We have trained the \ac{vqvae} on 1M and 2M tabular images, where $K = 8192$ for 1M, and $K = 16384$ for 2M.

\textbf{Image patches.}
Given an input tabular image $I \in \mathbb{R}^{H \times W \times C}$, the linear projection divides $I$ into a sequence of flattened 2D patches $I_p \in \mathbb{R}^{N \times (P^2 \cdot C)}$, where $C$ is the number of channels, $(P, P)$ is the size of each image patch, and $N = HW/P^2$ is the number of patches.
It is implemented by a kernel $P \times P$, stride $P$ convolution. 
We set $P=16$ and $I \in \mathbb{R}^{448 \times 448 \times 3}$, thus the sequence length of the image patches is $28 \times 28$. 
Approximately 40\% of the sequence is replaced with a masked token and the pretraining objective is to maximize the log-likelihood of the visual tokens of the masked region given the unmasked region. 
The image tokenizer's codebook provides the groundtruth visual tokens. 

The pretraining task is inspired by the success of masked language~\cite{devlin2018bert} and natural image~\cite{bao2021beit} modeling, but our work differs in the following ways:

(1) Table incorporates both vision and language representation of data, presenting concise human language with implicit conventions. 
Models for visual language understanding are required to read with high fidelity while also building rich high-level representation, relying on signals from both vision and language. 
In contrast, our work first explores the feasibility of self-supervised learning only on images for visual language tasks, using tables as an example. 

(2) A domain shift from semantic-rich natural images to predominantly black texts on white background tabular images poses a difficult optimization problem. 
It appears that tabular images are mainly text and lines separating rows and columns, so the visual codebook can quickly exhaust the table patterns with minimal tokens. 
Contrarily, the training of \ac{vqvae} can easily diverge. 
We discover that training stability is achieved by increasing the total number of tokens or introducing tabular images with colorful backgrounds. 
Sec.~\ref{sec: codebook} visualizes the semantics of the trained visual codebook and provides an explanation for the substantial number of tokens required.
\subsection{Unified Finetuning Framework}
\label{sec: method-unify}

Prior work employed a task-specific decoder for each task, where
$\mathbf{S}$ was predicted by an image-to-text pipeline and $\mathbf{B}$ was predicted by a detection head, \eg, Faster R-CNN~\cite{smock2022pubtables} or DETR~\cite{nassar2022tableformer}. 
The assumption on predicting $\mathbf{C}$ varies: some assume a \ac{pdf} is always accompanying the tabular image~\cite{nassar2022tableformer}, others rely on external text line detection and text recognition models~\cite{ye2021pingan, huang2023improving}. 
We aim to provide a unified task-agnostic training framework, where the input to our model is an image in the form of raw pixels only and the output is text in the form of token sequences. 
This setting is also generalizable to \ac{pdf} input modality as we can simply take a screenshot of the \ac{pdf}. 
The framework also enables us to leverage the visual encoder pretrained on the unannotated tabular images. 

\textbf{Table structure $\mathbf{S}$ }.
Predicting the table structure $\mathbf{S}$ already fits our training framework as $\mathbf{S}$ is defined by discrete HTML table tags. 
For non-spanning cells, we use \texttt{<td></td>} and \texttt{<td>[]</td>} to denote empty and non-empty cells. 
For spanning cells, \texttt{<td} marks the beginning, and \texttt{></td>} and \texttt{>[]</td>} marks the ending of empty and non-empty cells. 
The specific tokens for spanning cells are \texttt{rowspan="n"} and \texttt{colspan="n"}. 
We use $n \in [2, 19]$ as that covers most of the tables in practice. 
Apart from the data cell tags, the vocabulary also contains the following tags that define a table: \texttt{<thead>},  \texttt{<tbody>},  \texttt{<tr>}, and their corresponding closing tags. 

\textbf{Table cell bbox $\mathbf{B}$}.
Each cell \ac{bbox} $\mathbf{b}$ are four continuous coordinates, which are not naturally expressed as discrete tokens. 
Inspired by Pix2Seq~\cite{chen2021pix2seq}, we discretize the coordinate into an integer between 0 and image size.
The two directions of an image share the same vocabulary. 
Since we need to predict all bboxes within a tabular image, each quantized \ac{bbox} is concatenated together in reading order: from left to right and top to bottom.
This formulation completes the quantization and serialization of all bboxes into a sequence of discrete tokens. 
At inference time, we de-serialize the predicted sequence into groups of four tokens. 

\textbf{Table cell content $\mathbf{C}$}.
After predicting all \ac{bbox} coordinates, we only need to perform \ac{ocr} on the image region within each \ac{bbox}. 
Note that each cell is defined by a single bbox, thus the cell content can have single or multiple lines of text. 
In the training stage, the model is trained on a mixture of single line and multi-line dataset. 
At inference time, we parse all text simultaneously as each cell \ac{bbox} is independent. 
Finally, we insert the cell content back into the non-empty cells \texttt{<td>[]</td>} or \texttt{>[]</td>} as the reading order is already preserved in both $\textbf{S}$ and $\textbf{B}$.
We use WordPiece tokenizer~\cite{wu2016google} with charater-level granularity since \ac{ocr} requires the model to read instead of understanding the semantics, which significantly reduces the total vocabulary size to less than 6k.

Up till now, we have completely digitalized a tabular image into HTML with a unified image-to-text framework. 
Note that all visual encoders are initialized from \ac{ssp}. 
For cell content and cell bbox, there is an alternative solution that we first generate all the cell content within a table, and then predict the cell bbox via prompting the model with cell content. 
However, we find it hard for the model to predict all the cell content first as all tabular images are rescaled to a fixed size during augmentation, and such rescaling leads to texts in various aspect ratios.
Thus, we do not use this solution and instead generate all cell \ac{bbox} first. 

\textbf{Training Objective.}
Since all task outputs have been formulated into a sequence of discrete tokens, the objective function is simply the maximum likelihood of tokens conditioned on pixel inputs and the preceding tokens. 
Denote the probability of the $i$th step prediction $p(t_i | I, t_{1: i - 1}; \theta)$, we directly maximize the correct structure prediction by using the following formulation:
\begin{equation}
    \theta^\ast = \argmax_\theta \sum_{(I, T)} \log p(T | I; \theta),
\end{equation}
where $\theta$ are model parameters.



\begin{table}[!tbp]
\centering
\small
\caption{\method{} outperforms prior methods and achieves \ac{sota} on four out of five largest publicly available \ac{tsr} datasets across all available tasks. Our method is trained with a task-agnostic language modeling loss and does not rely on external \ac{pdf} for text extraction and \ac{bbox} post-processing. 
}
\begin{tabular}{l@{\hspace{3mm}}c@{\hspace{1mm}}c@{\hspace{2mm}}c@{\hspace{1mm}}c@{\hspace{1mm}}c@{\hspace{2mm}}c@{\hspace{2mm}}c@{\hspace{1mm}}c@{\hspace{2mm}}c@{\hspace{1mm}}c}
\toprule
    & \multicolumn{2}{c}{\acs{icdar19}} & \multicolumn{3}{c}{PubTabNet} & FinTabNet & \multicolumn{2}{c}{SynthTabNet} & \multicolumn{2}{c}{PubTables-1M} \\
    & IoU 0.6 & WAvg. F1 & $\text{AP}_{50}$ & S-TEDS & TEDS & S-TEDS & $\text{AP}_{50}$ & S-TEDS & $\text{AP}_{50}$ & $\text{AP}_{75}$\\ 
\cmidrule(r){2-3}\cmidrule(r){4-6}\cmidrule(r){7-7}\cmidrule(r){8-9}\cmidrule(r){10-11}
    \multirow{2}{*}{\ac{sota}} 
    & GTE   & GTE   & VAST  & VAST  & VAST  & VAST  & TableFormer & DRCC  &   DETR & DETR \\
    & 38.50 & 24.80 & 94.50 & 97.23 & 96.31 & 98.63 & 87.70       & 98.70 &  \textbf{97.10} & \textbf{94.80} \\
\midrule
    \multicolumn{3}{l}{\textit{\method}} \\
    Base  & 54.97 & 40.15 & 97.94 & 95.63 & 94.78 & 97.19 & 98.99 & 98.97 & 94.48 & 88.64 \\
    Large & \textbf{58.10} & \textbf{42.62} & \textbf{98.43} & \textbf{97.89} & \textbf{96.50} & \textbf{98.89} & \textbf{99.00} & \textbf{99.39} & 95.68 & 93.28 \\
\bottomrule
\end{tabular}
\label{tab:sota}
\end{table}

\section{Experiments}
\label{sec: exp}

\subsection{Implementation}

\textbf{Architecture.} 
We have trained two model variants: (1) a \textit{base} model with 30M parameters including 4 encoder layers, 8 attention heads with a hidden size of 512, (2) a \textit{large} model with 125M parameters including 12 encoder layers, 12 attention heads with a hidden size of 768. 
Both base and large models have a task decoder of 4 decoder layers. 
The maximum token sequence length is 512 for table structure, 1024 for cell bbox, and 200 for cell content, as we find such settings satisfy most tables.

\textbf{Training and inference}
We have pretrained the \ac{vqvae} on 1M and 2M tabular images. 
The 1M \ac{vqvae} is trained on PubTabNet~\cite{zhong2020image} and SynthTabNet~\cite{nassar2022tableformer}, and the extra 1M datasets for training 2M \ac{vqvae} are PubTables-1M~\cite{smock2022pubtables} and TableBank~\cite{li2020tablebank}. 
In Sec.~\ref{sec: dataset}, we present the finetuning results of 2M \ac{vqvae} for comparing with \ac{sota} methods. 
In Sec.~\ref{sec: ablation-ssp}, we ablate the effectiveness and scalability of \ac{ssp} on both 1M and 2M \ac{vqvae}. 
All models are trained with the AdamW optimizer~\cite{loshchilov2017decoupled}. 
We employ a cosine learning rate scheduler with a linear warmup. 
All models are trained for 24 epochs for a fair comparison. 
We apply teacher forcing during training and employ greedy decoding at inference time. 


\subsection{Evaluation Metrics}
\textbf{\Ac{car}} was first proposed in ICDAR2013 competition~\cite{gobel2013icdar} and improved by ICDAR2019 competition~\cite{gao2019icdar}.
It aligns the the predicted \ac{bbox} with the groundtruth \ac{bbox} for each table cell based on \ac{iou} and generates a list of adjacency relations between a non-empty cell and its nearest horizontal and vertical neighbors. 
The precision, recall, and F1 score are computed based on this converted 1-D adjacency relation list. 

\textbf{COCO \ac{ap}}~\cite{lin2014microsoft} is a widely used metric for generic object detection, which has been reported in other work for evaluating table cell detection. 
We use the COCO evaluation toolkit\footnote{https://github.com/cocodataset/cocoapi} and report \ac{map}, $\text{AP}_{50}$, and $\text{AP}_{75}$. 

\textbf{\Ac{teds}} was created by PubTabNet~\cite{zhong2020image} to robustify the \ac{car} metric against cell shift perturbation and cell content perturbation. 
\Ac{teds} converts the table HTML code into a tree structure and measures the edit distance between the prediction $T_{pred}$ and the groundtruth $T_{gt}$. 
A shorter edit distance indicates a higher degree of similarity, leading to a higher \ac{teds} score.
\Ac{teds} measures both the table structure and table cell content, and we use S-\ac{teds}~\cite{huang2023improving} when only the table structure is considered. 

\subsection{Results on Datasets}
\label{sec: dataset}

We evaluate \method{} on five of the largest publicly available \ac{tsr} datasets as shown in Table~\ref{tab:sota}.
Comparing \method{} with the prior \ac{sota} on each dataset across all available tasks, we achieve new \ac{sota} on four out of the five datasets even without training with task-specific loss~\cite{huang2023improving} or relying on external \ac{pdf} for text extraction and \ac{bbox} post-processing~\cite{nassar2022tableformer}.
Below is an introduction of each dataset and comparisons with previous methods. 

\textbf{\ac{icdar19}}~\cite{gao2019icdar} was originated from ICDAR 2019 table competition.
The dataset has two subsets for \ac{tsr}, archival and modern, and only the modern subset has table cell content \ac{bbox} annotations. 
The competition computes \ac{car} F1 score at $\text{\acs{iou}} \in [0.6, 0.7, 0.8, 0.9]$ and ranks method by \ac{wf1}:
\begin{equation}
    \text{\acs{wf1}} = \frac{\sum_{i = 1}^4 \text{\acs{iou}}_i \times \text{F1@\acs{iou}}_i}{\sum_{i = 1}^4 \text{\acs{iou}}_i}
\end{equation}
We evalute on all 100 test tables from the modern subset and report both \ac{wf1} and \ac{iou} 0.6 as in other work.  
Our method significantly improves the previous \ac{sota} GTE~\cite{zheng2021global} by a large margin. 

\textbf{PubTabNet}~\cite{zhong2020image} contains 509k images of heterogeneous tables extracted from the medical scientific articles.
It is the first large-scale \ac{tsr} dataset that provides annotations (in HTML format) of table cell \ac{bbox}, table structure, and table cell content. 
PubTabNet measures the table cell \ac{bbox} by COCO $\text{AP}_{50}$, table structure by S-\ac{teds}, and full table including both structure and cell content by \ac{teds}.
The authors of PubTabNet also developed the EDD model, which consisted of a \ac{cnn} encoder and dual \ac{rnn} decoders for predicting table structure and cell content, respectively. 
TableFormer~\cite{nassar2022tableformer} improved EDD by replacing the cell content decoder with a cell \ac{bbox} decoder and extracted all contents from the \ac{pdf} corresponding to the tabular image. 
VAST~\cite{huang2023improving} added an auxiliary visual-alignment loss while training the cell \ac{bbox} decoder and achieved previous \ac{sota} on all three metrics. 
Our unified training framework with \ac{ssp} achieves the new \ac{sota} on all tasks even without leveraging any external \ac{pdf} as \ac{pdf} corresponding to the tabular image may not always exist. 
Specifically, both \method{}-base and \method{}-large outperform VAST on $\text{AP}_{50}$ by more than 3 \ac{pp}, which confirms the effectiveness of converting the cell \ac{bbox} detection to language modeling. 

\textbf{FinTabNet}~\cite{zheng2021global} is a dataset containing 113k tables from the annual reports of the S\&P 500 companies in \ac{pdf} format. 
The major challenge of this dataset is that financial tables largely differ from scientific and government document tables in that the former has fewer graphical lines, larger gaps within each table, and more color variations. 
Thus, FinTabNet mainly evaluates table structure prediction accuracy, \ie, S-\ac{teds}.
VAST trained the model with an auxiliary supervised signal and achieved the previous \ac{sota}. 
We achieve the new \ac{sota} by leveraging \ac{ssp} and finetuning without a task-specific loss objective. 
The performance gain from base to large shows that our method scales with model parameters. 

\textbf{SynthTabNet}~\cite{nassar2022tableformer} is a large-scale synthetically generated dataset that offers control over 1) dataset size, 2) table structure, 3) table style, and 4) content type. 
The dataset aims to overcome the limitations of PubTabNet and FinTabNet, which are skewed table distributions towards simpler tables, limited variance in appearance styles, and restricted cell content domains. 
Thus, SynthTabNet is organized into 4 subsets of 150k tables (600k in total), namely Finance, PubTabNet, Marketing, and Sparse.
The first two mimic the appearance of FinTabNet and PubTabNet but encompass more complex table structures. 
Marketing adopts a colorful appearance with high contrast that resembles real-world marketing tables, and Sparse contains tables with sparse content. 
The authors of SynthTabNet propose TableFormer as a baseline. 
TableFormer predicts table structure and cell \ac{bbox} and relies on external \ac{pdf} to extract cell content, so $\text{AP}_{50}$ and S-\ac{teds} are the evaluation metrics. 
Both our base and large models outperform previous \ac{sota} on $\text{AP}_{50}$ and S-\ac{teds}. 
Since SynthTabNet is a comprehensive dataset that can thoroughly evaluate the model under different table configurations, we use it for ablations and present results on four subsets separately in Sec.~\ref{sec: analysis}.

\textbf{PubTables-1M}~\cite{smock2022pubtables} aims to overcome the groundtruth inconsistency observed in prior datasets using a new canonicalization procedure. 
The dataset has 947k tables annotated with \ac{bbox} and text within each \ac{bbox}. 
The dataset differs from previous datasets that the \ac{bbox} is word-wise instead of cell-wise, thus each cell can have more than one \ac{bbox}. 
Besides, since all annotations are in \ac{bbox} format and no table structure labels, \eg, HTML, are provided, the baseline DETR trained by the dataset creators report their performance in detection metrics, \eg, $\text{AP}_{50}$ and $\text{AP}_{75}$. 
\method{} also achieves competitive results compared with the baseline DETR. 
Visualizing the \ac{bbox} predictions  shows that our model  predicts more bboxes (longer sequence) than the groundtruth, 
motivating us to delve deep into the dataset annotations and discover previously unacknowledged inconsistencies in table annotations. 
We conjecture the \ac{sota} performance of DETR model on PubTables-1M may be due to overfitting to the training set. 
Sec.~\ref{sec: annotation} describes the details of these dataset annotation issues.

\subsection{Using \method{} in Practice}
We have also finetuned our \method{}-large across multiple datasets and released a Jupyter Notebook as a demo of our inference pipeline. 
A user can simply pass a table screenshot through our notebook and obtain a fully digitalized HTML table. 
The table structure is trained across PubTabNet, FinTabNet, and SynthTabNet, the cell \ac{bbox} is trained across PubTabNet and SynthTabNet, and the cell content is trained across PubTabNet, SynthTabNet, and PubTables-1M. 
We have provided a public API hosted on HuggingFace to facilitate the access to our UniTable: 
\url{https://poloclub.github.io/magic-table/}.
Appendix~\ref{apppx: demo} visualizes several table examples in practice. 




\section{Deeper Dive into \method{}: Ablation and Analysis}
\label{sec: analysis}

\subsection{Effectiveness and Scalability of SSP}
\label{sec: ablation-ssp}


\begin{table}
\small
\centering
\caption{Effectiveness and scalability of \ac{ssp} on all four subsets of SynthTabNet. 
Without \ac{ssp}, the model performance suffers, and increasing the model complexity from base to large barely improves the performance.
Pretraining the visual encoder on 1M tabular images provides an average increase of 14.40 \ac{pp}. 
Pretraining on 2M images continues to increase the performance by 0.74 \ac{pp}.
Here we present S-\ac{teds} of the table structure prediction, and the same trend also applies to other tasks as elaborated in Appendix~\ref{appx: ablation-ssp}. 
}
\begin{tabular}
{@{}l@{\hspace{2mm}}c@{\hspace{1mm}}c@{\hspace{2mm}}c@{\hspace{1mm}}c@{\hspace{2mm}}c@{\hspace{1mm}}c@{\hspace{2mm}}c@{\hspace{1mm}}c@{}}

\toprule
    & \multicolumn{2}{c}{Finance} & \multicolumn{2}{c}{PubTabNet} & \multicolumn{2}{c}{Marketing} & \multicolumn{2}{c}{Sparse} \\
    & Base & Large & Base & Large & Base & Large & Base & Large \\
\cmidrule(r){2-3}\cmidrule(r){4-5}\cmidrule(r){6-7}\cmidrule(r){8-9}
    No \acs{ssp} & 88.95 & 90.75 & 89.10 & 91.67 & 68.05 & 70.60 & 85.50 & 87.72 \\
    \acs{ssp} 1M & 98.73 & 99.56 & 99.02 & 99.55 & 95.14 & 99.05 & 97.20 & 99.29 \\
    \acs{ssp} 2M & 99.41 & \textbf{99.58} & 99.44 & \textbf{99.56} & 98.35 & \textbf{99.08} & 98.69 & \textbf{99.34} \\

\bottomrule
\end{tabular}
\label{tab:ablation-ssp}
\end{table}

We ablate the effectiveness and scalability of \ac{ssp} on all four subsets of SynthTabNet since SynthTabNet is a large-scale dataset that can comprehensively evaluate the model under different table configurations. 
Table~\ref{tab:ablation-ssp} presents the table structure task, and the same trend also applies to other tasks as elaborated in Appendix~\ref{appx: ablation-ssp}. 
Comparing row ``No SSP'' and ``SSP 1M'' or ``SSP 2M'', both the base and the large models have benefited significantly from the \ac{ssp}. 
Specifically, Marketing and Sparse are two of the most challenging subsets. 
Marketing has a colorful background with high-contrast texts, and Sparse contains tables with sparse content. 
These variations make it challenging to accurately predict the HTML table structure tags and \ac{bbox} surrounding the cell content. 
Without \ac{ssp}, the model performance suffers, and increasing the model complexity from base to large barely improves the performance. 
Instead, after pretraining the visual encoder on 1M tabular images, both the base and the large models have an average increase of 14.40 \ac{pp}. 
Note the performance also scales along with the pretraining dataset size. 
When the pretraining dataset increases from 1M to 2M, the average performance continues to increase by 0.74 \ac{pp}. 
Finally, all large models have an average performance gain of 1.51 \ac{pp} over the base models. 


\subsection{Generalization of the Unified Training Objective}
\label{sec: ablation-unify}

\begin{table}[!htbp]
\small
\centering
\caption{\method{}’s unified training objective applies to both linear projection Transformer and hybrid CNN-Transformer architectures conventionally used in \ac{tsr}. 
Results on all four subsets of the SynthTabNet for table structure prediction evaluated with S-\ac{teds}, and the same conclusion also applies to other tasks as elaborated in Appendix~\ref{appx: ablation-unify}. }
\begin{tabular}
{lcccc}
\toprule
    Model & Finance & PubTabNet & Marketing & Sparse \\
\midrule
    Base  & 98.63 & 98.80 & 97.16 & 95.30 \\
    Large & 99.44 & 99.44 & 98.71 & 98.64 \\
\bottomrule
\end{tabular}
\label{tab:ablation-unify}
\end{table}

We showcase that our unified training objective, language modeling, not only works on models with linear projection Transformer, but also works with the hybrid \ac{cnn}-Transformer architecture used in the \ac{tsr} literature. 
Table~\ref{tab:ablation-unify} presents the results on all four subsets of the SynthTabNet similar to the settings in Sec.~\ref{sec: ablation-ssp}. 
The main difference is that the linear projection in the visual encoder is replaced by a ResNet-18 \ac{cnn} backbone. 
Such a modification leads to an increase of 12M parameters. 
The performance of this hybrid \ac{cnn}-Transformer architecture is roughly on par with the linear projection model initialized from the \ac{ssp} 1M tabular images, which shows that our training objective is agnostic to the choice of architecture. 
The significant performance gap between hybrid \ac{cnn}-Transformer and linear projection Transformer trained from scratch verifies the observation identified in the previous literature~\cite{peng2023high}. 
Though hybrid \ac{cnn}-Transformer has also achieved competitive results, we still recommend using the linear projection Transformer because of 1) capability of leveraging the power of \ac{ssp}, 2) architectural compliance with \ac{vlm} in natural image domain, and 3) the performance of hybrid \ac{cnn}-Transformer is still worse than the \ac{ssp} on 2M tabular images even with more total parameters.
Though prior work has started to propose \ac{ssp} for hybrid \ac{cnn}-Transformer, this direction is still in its early stage as it is challenging in heavy computational cost and pretraining-finetuning discrepancy~\cite{gao2022convmae}.
Moreover, most work only demonstrates the capability in discriminative tasks rather than the generative tasks used in the \ac{tsr} domain. 
Besides, existing \ac{sota} \acp{vlm}, \eg, BLIP-2~\cite{li2023blip} and LLaVA~\cite{liu2023visual}, still employ the linear projection and leave the Transformer to learn the interactions between different patches. 
As we believe the compliance of architecture is an important step towards generic \acp{vlm}, our success in replacing the \ac{cnn} backbone with linear projection and leveraging a task-agnostic language modeling loss is a cornerstone of incorporating \ac{tsr} in modern \ac{vlm} training. 

\subsection{Why does SSP work?}
\label{sec: codebook}

\begin{figure}
\centering
\begin{subfigure}[t]{0.3\textwidth}
    \centering
    \includegraphics[width=\textwidth]{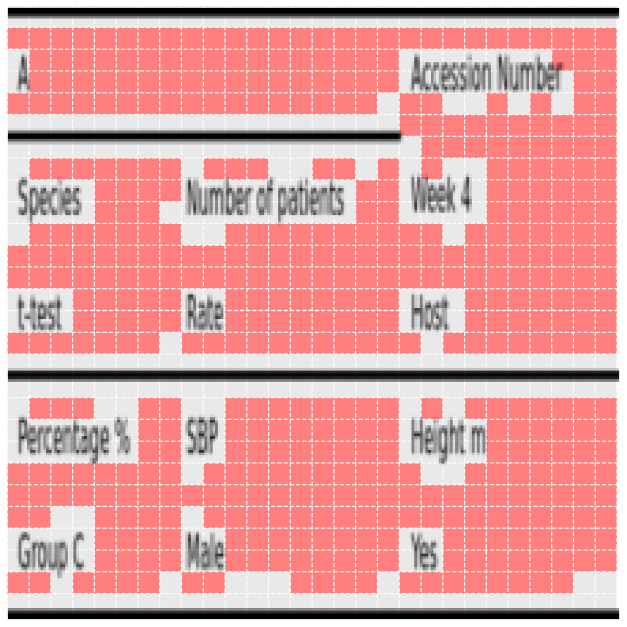}
    \caption{\textit{Background} patches (red)}
    \label{fig:codebook-a}
\end{subfigure}
\hspace{5mm}
\begin{subfigure}[t]{0.3\textwidth}
    \centering
    \includegraphics[width=\textwidth]{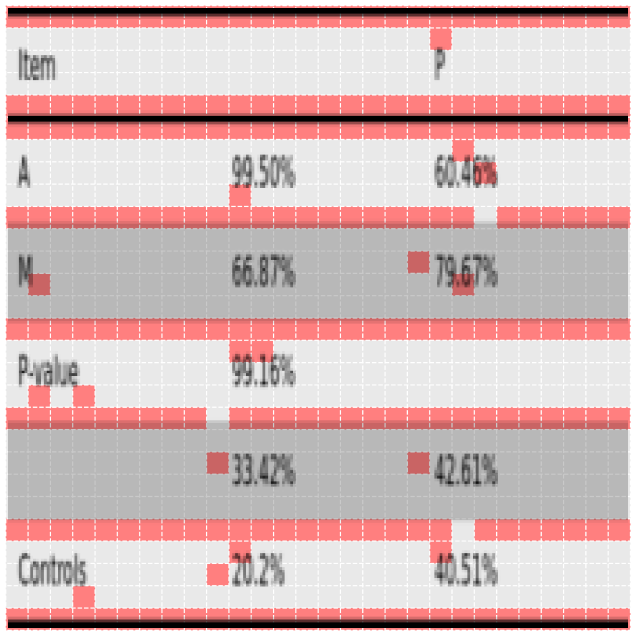}
    \caption{Patches for \textit{separations}}
    \label{fig:codebook-b}
\end{subfigure}
\caption{
The key reason that \ac{ssp} works is because each tokens have visual semantics and the codebook shows a fine-grained categorization to represent the implicit conventions in the table.  
The codebook used in \ac{ssp} has learned to represent abstract concepts by using different groups of tokens to represent different concepts: (a) empty background and (b) separations within a table. 
Red highlights the token indices under investigation.
Appendix~\ref{appx: codebook} provides a zoomed-in version of these images labeled with token indices.
We lay the color patches over the original tabular image to present the selected token indices from the 2M \ac{vqvae}. 
}
\label{fig:codebook}
\end{figure}

During \ac{ssp}, the visual encoder is trained to fill in the masked tabular image by selecting visual tokens from the \ac{vqvae} codebook. 
Thus, we visualize the selected token indices and analyze whether the tokens have visual semantics. 
Fig.~\ref{fig:codebook} presents the original table overlaid by the selected token indices from the 2M \ac{vqvae} codebook. 
We highlight the token indices under investigation in red. 
Appendix~\ref{appx: codebook} provides a zoomed-in version of these images labeled with token indices.
The input image size is $448 \times 448$ and the patch size is $16 \times 16$, so each image has $28 \times 28$ tokens chosen from the 16384 entries in the codebook. 
First, we highlight the tokens representing the blank  background (Fig~\ref{fig:codebook}a), which covers most of the empty space in the table; 
leaving only texts and separation lines nonhighlited.
Then, we highlight the tokens representing the separation lines (Fig~\ref{fig:codebook}b), showing that they well-separate the header row, and the shaded and non-shaded rows. 
Taking a closer look at the token indices in Fig.~\ref{fig:appx-codebook-space} and \ref{fig:appx-codebook-separation} in Appendix~\ref{appx: codebook}, we find the codebook has learned to represent an abstract concept by assigning multiple tokens, where each reflects a different scenario.
For example, token ``111'' represents the separation above the gray color shading and token ``964'' represents the one below, and token ``15282'' represents the separation above a bold horizontal line and ``10807'' represents the one below. 
A certain type of separation also has multiple choices depending on the portion of the line within the patch, \eg, token ``14181'', ``8714'', and ``10807'' all representing the ``below a bold horizontal line'', but the differences lie in the line thickness and amount of black inside the image patch. 
Such a fine-grained categorization also explains why the codebook needs so many tokens to represent the implicit conventions created by humans in the table.







\section{Conclusion}
\vspace{-2mm}
We present \method{}, a training framework that unifies both the training paradigm and training objective of \ac{tsr}.
Its training paradigm combines the simplicity of purely pixel-level inputs with the effectiveness and scalability empowered by \ac{ssp} from diverse unannotated tabular images.
Our framework unifies the training objectives of all three \ac{tsr} tasks, extracting table structure, cell content, and cell \ac{bbox}, into a unified task-agnostic training objective: language modeling. 
Extensive quantitative and qualitative analyses highlights \method{}'s \ac{sota} performance on four of the largest \ac{tsr} datasets
To promote reproducible research, enhance transparency, and SOTA innovations, we open-source our code and release the first-of-its-kind Jupyter Notebook of the whole inference pipeline, fine-tuned across multiple \ac{tsr} datasets, supporting all three \ac{tsr} tasks.

\clearpage
\bibliographystyle{plainnat}
\bibliography{ref}

\clearpage
\appendix
\section{Detail comparisons with milestone VLMs}
\label{appx: vlm}


\begin{table}[!htbp]
\small
\centering
\caption{\method{} outperforms GPT-4V and LLaVA-v1.6 by a huge margin for table recognition.}
\begin{tabular}{@{}l@{\hspace{2mm}}r@{\hspace{1.5mm}}r@{\hspace{1.5mm}}r@{\hspace{2mm}}r@{\hspace{1.5mm}}r@{\hspace{1.5mm}}r@{\hspace{2mm}}r@{\hspace{1.5mm}}r@{\hspace{1.5mm}}r@{\hspace{2mm}}r@{\hspace{1.5mm}}r@{\hspace{1.5mm}}r@{}}
\toprule
Model & \multicolumn{3}{c}{Finance} & \multicolumn{3}{c}{PubTabNet} & \multicolumn{3}{c}{Marketing} & \multicolumn{3}{c}{Sparse}  \\
\midrule
Samples & 100 & 300 & 500 & 100 & 300 & 500 & 100 & 300 & 500 & 100 & 300 & 500 \\
\midrule
LLaVA-v1.6-vicuna-7b & 37.72 & 35.57 & 34.16 & 41.66 & 42.19 & 41.80 & 36.54 & 34.34 & 34.14 & 36.95 & 37.42 & 38.49 \\
LLaVA-v1.6-mistral-7b & 38.40 & 39.76 & 38.07 & 39.16 & 37.79 & 38.96 & 34.16 & 33.20 & 32.25 & 40.28 & 35.97 & 37.06 \\
LLaVA-v1.6-vicuna-13b & 43.50 & 45.13 & 43.95 & 48.17 & 49.05 & 49.68 & 41.37 & 42.00 & 41.40 & 46.38 & 46.04 & 47.21 \\
LLaVA-v1.6-34b & 40.86 & 42.96 & 40.69 & 46.30 & 47.99 & 49.30 & 35.08 & 35.75 & 33.99 & 39.13 & 39.60 & 39.85 \\
GPT-4-1106-vision-pre. & 64.05 & 66.93 & 66.13 & 69.26 & 69.86 & 69.62 & 58.51 & 60.03 & 59.13 & 55.64 & 57.02 & 56.99 \\
UniTable Large & \textbf{99.63} & \textbf{99.56} & \textbf{99.57} & \textbf{99.54} & \textbf{99.54} & \textbf{99.55} & \textbf{99.11} & \textbf{99.10} & \textbf{99.10} & \textbf{99.34} & \textbf{99.35} & \textbf{99.34} \\
\bottomrule
\end{tabular}
\label{tab:vlm_comparison}
\end{table}

Table~\ref{tab:vlm_comparison} shows detailed comparisons (S-TEDS score) with GPT-4V and the latest LLaVA v1.6. 
In summary, UniTable outperforms GPT-4V and LLaVA-v1.6 by a huge margin for table recognition.

\textbf{Dataset}: SynthTabNet~\cite{nassar2022tableformer}, is a large-scale table benchmark that offers control over dataset size, table structure, style, and content type. 
It has 4 subsets: Finance, PubTabNet, Marketing, Sparse. 
We incrementally evaluate more samples up to 500 on each subset because no further notable changes in the results are observed.

\textbf{Models}: We compare UniTable with GPT-4V and LLaVA. 
For GPT-4V, we use the gpt-4-1106-vision-preview. 
For LLaVA, we test on all 4 variants of its latest version, v1.6, provided by the official model weights on HuggingFace~\cite{liu2024visual}: vicuna-7b, mistral-7b, vicuna-13b, 34b. 
We use the same prompt as in the latest work that evaluates GPT-4V's OCR capabilities by Shi \etal, \cite{shi2023exploring}: ``Please read the table in this image and return an HTML-style reconstructed table in text, do not omit anything.'' 
The max generated sequence length is set to 1500 to cover the largest table in the test. 
\section{Using \method{} in Practice}
\label{apppx: demo}

\begin{figure}
\centering
\begin{subfigure}[b]{0.8\textwidth}
    \centering
    \includegraphics[width=\textwidth]{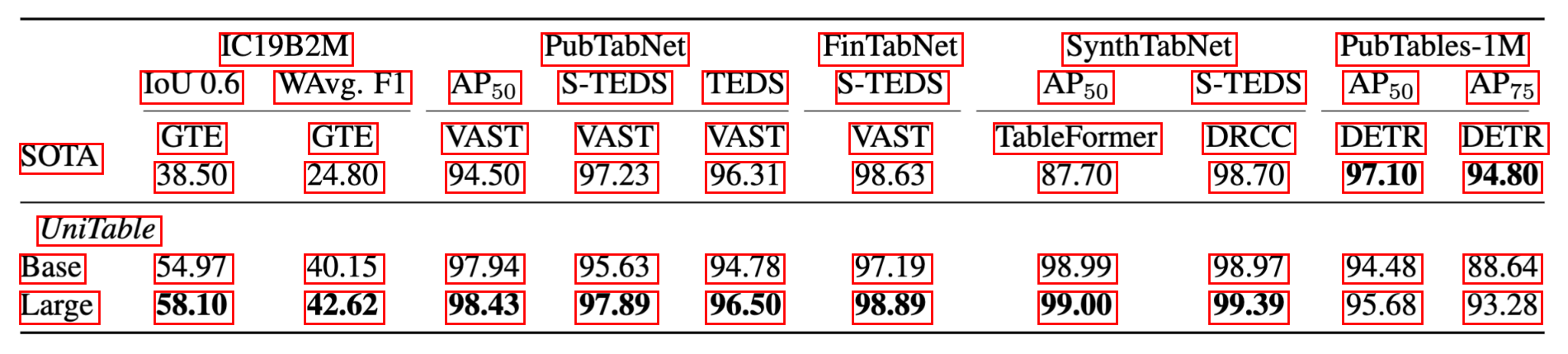}
    \caption{Table cell \ac{bbox} detections results on Table~\ref{tab:sota} in the main paper.}
    \label{fig:demo-a}
\end{subfigure}
\vskip\baselineskip
\begin{subfigure}[b]{0.8\textwidth}
    \centering
    \includegraphics[width=\textwidth]{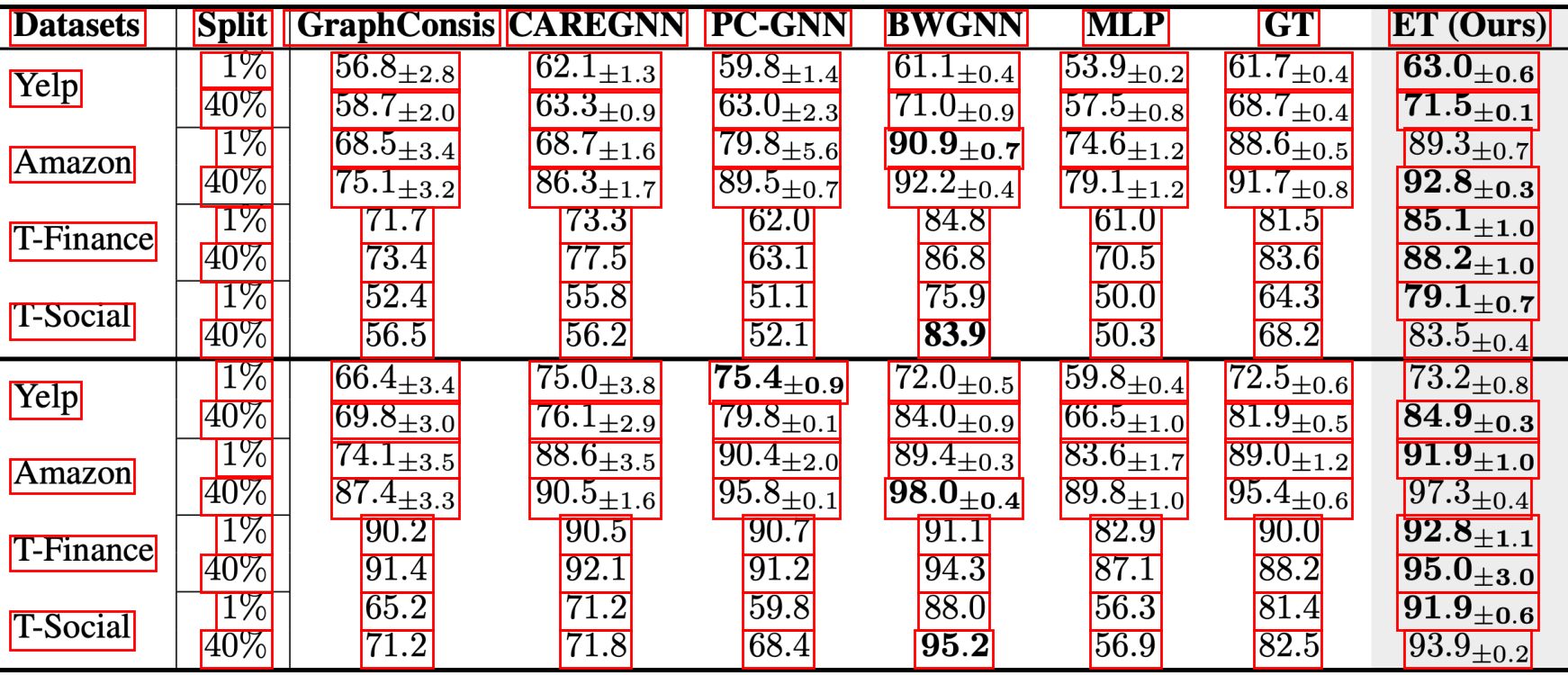}
    \caption{Table cell \ac{bbox} detections results on complex academic tables}
    \label{fig:demo-b}
\end{subfigure}
\vskip\baselineskip
\begin{subfigure}[b]{0.8\textwidth}
    \centering
    \includegraphics[width=\textwidth]{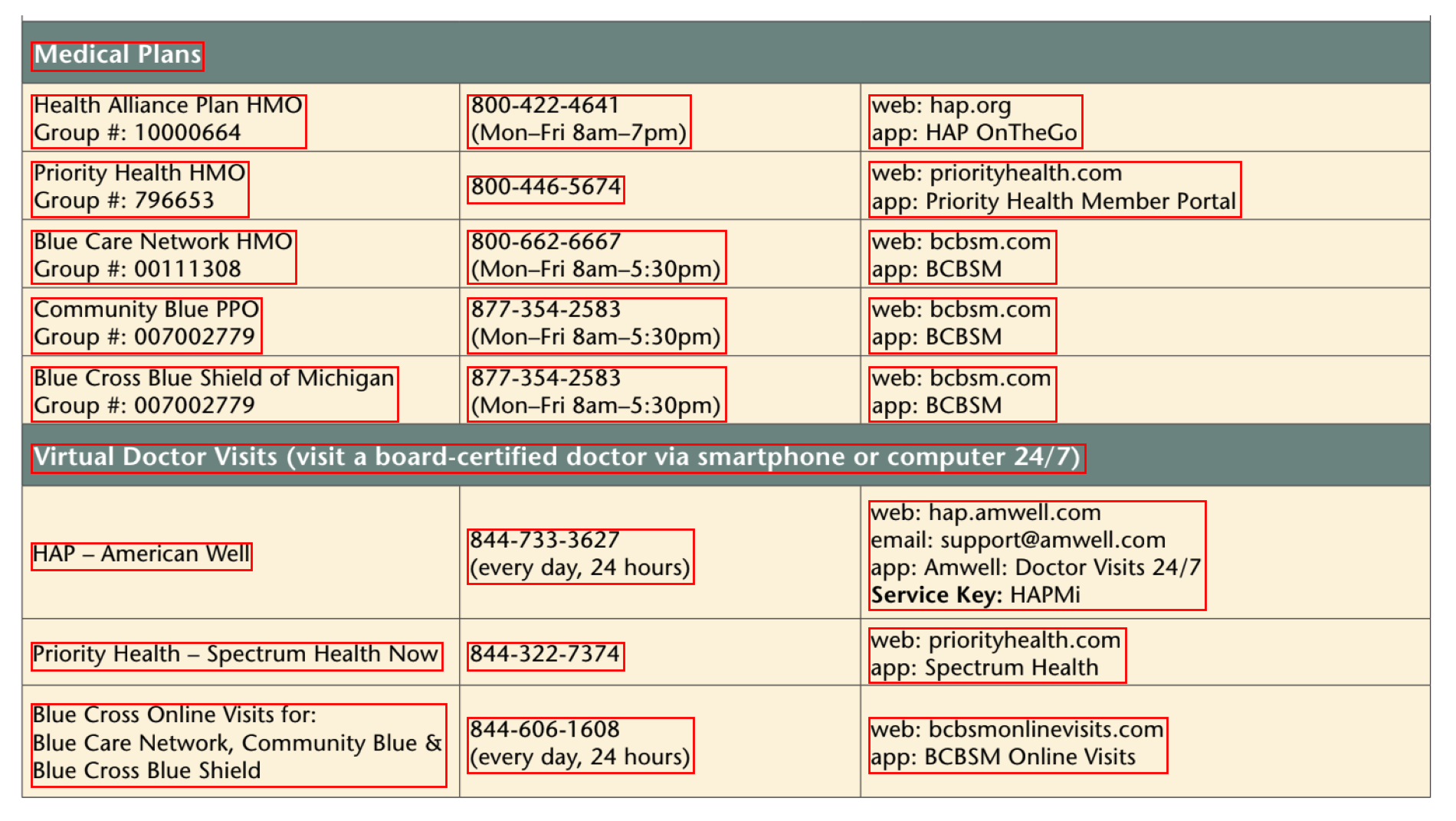}
    \caption{Table cell \ac{bbox} detections results on tables with colorful backgrounds and spanning headers.}
    \label{fig:demo-c}
\end{subfigure}
\caption{
Table cell \ac{bbox} detection results of \method{} on unannotated tables in practice.}
\label{fig:demo}
\end{figure}

A user can simply pass a table screenshot through our notebook and obtain a fully digitalized HTML table.
We visualize different types of table cell \ac{bbox} detection results of \method{} in Fig.~\ref{fig:demo}, and an example of a digitalized HTML table can be found in \url{https://anonymous.4open.science/r/icml-review/notebooks/full_pipeline.ipynb}.

\section{Effectiveness and Scalability of SSP}
\label{appx: ablation-ssp}

Table~\ref{tab:appx-ablation-ssp} demonstrates the effectiveness and scalability of \ac{ssp} on all four subsets of SynthTabNet. 
Without \ac{ssp}, the model performance suffers, and increasing the model complexity from base to large barely improves the performance.
Pretraining the visual encoder on 1M tabular images provides a significant improvement, and pretraining on 2M images continues to increase the performance on all \ac{tsr} tasks.

\section{Generalization of the Unified Training Objective}
\label{appx: ablation-unify}

\method{}’s unified training objective applies to both linear projection Transformer and hybrid CNN-Transformer architectures conventionally used in \ac{tsr}. 
Table~\ref{tab:appx-ablation-unify} shows results on all four subsets of the SynthTabNet for table structure prediction evaluated with S-\ac{teds} and table cell \ac{bbox} detection evaluated with \ac{map}.
\section{Visualization of the Visual Codebook}
\label{appx: codebook}

\begin{figure}[ht]
\begin{center}
\centerline{\includegraphics[width=\columnwidth]{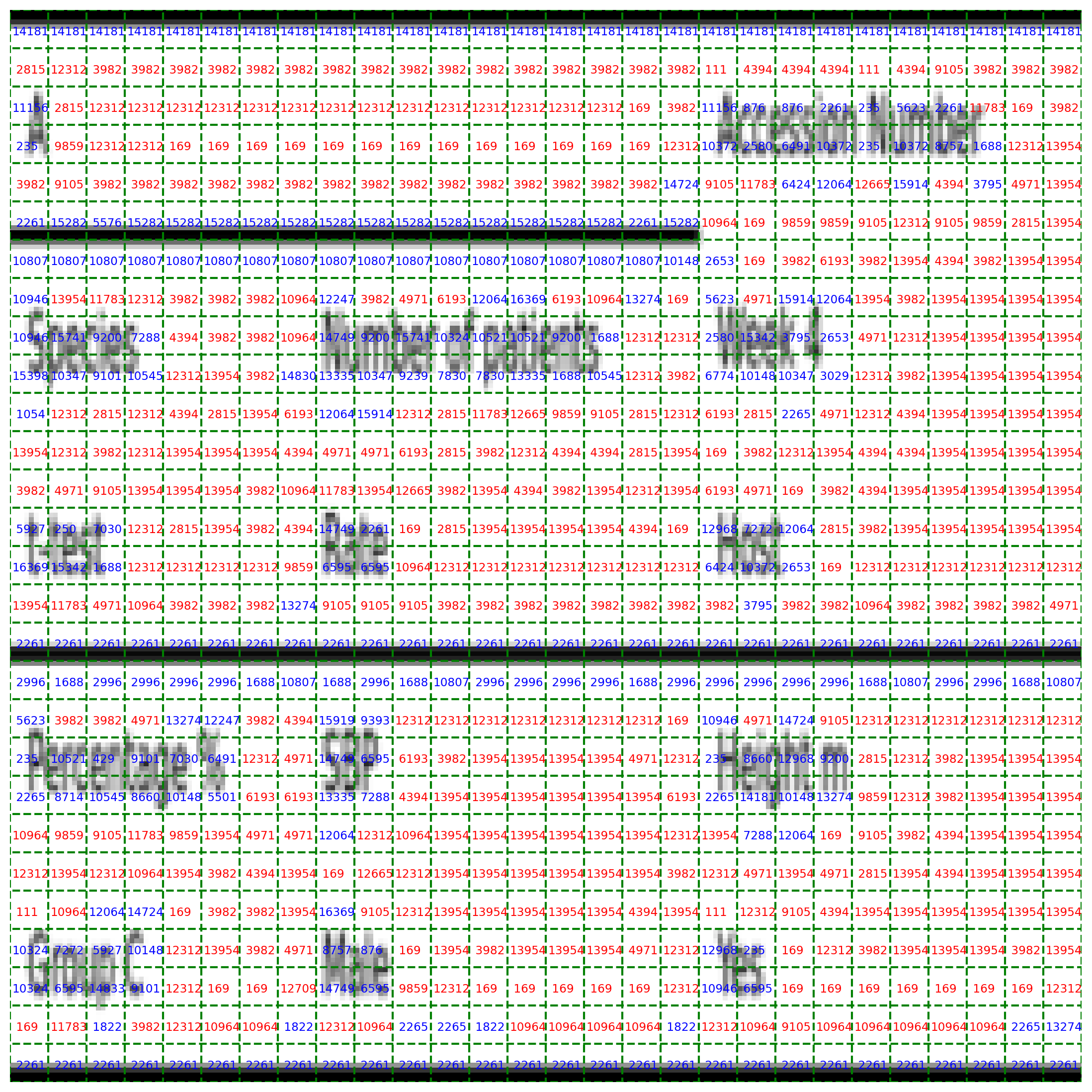}}
\caption{A zoomed-in version of the token indices from the 2M \ac{vqvae}. 
The codebook used in \ac{ssp} has learned to represent abstract concepts by using different groups of tokens to represent empty white backgrounds within a table.
Red highlights the token indices representing the concept of ``empty white backgrounds''.}
\label{fig:appx-codebook-space}
\end{center}
\end{figure}

\begin{figure}[ht]
\begin{center}
\centerline{\includegraphics[width=\columnwidth]{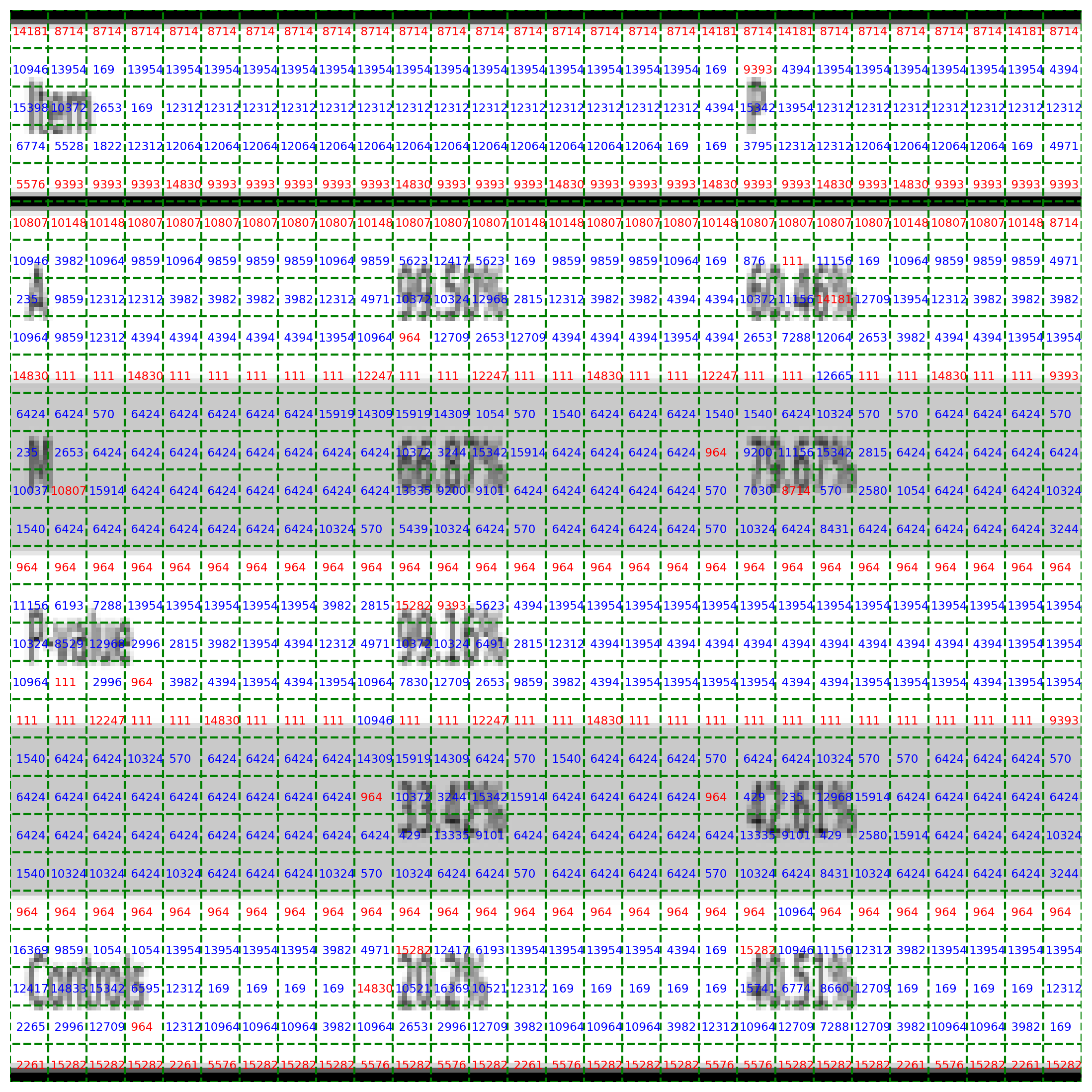}}
\caption{A zoomed-in version of the token indices from the 2M \ac{vqvae}. 
The codebook used in \ac{ssp} has learned to represent abstract concepts by using different groups of tokens to represent separations within a table.
Red highlights the token indices representing the concept of ``separations''.}
\label{fig:appx-codebook-separation}
\end{center}
\end{figure}

Fig.~\ref{fig:appx-codebook-space} and \ref{fig:appx-codebook-separation} present the original table overlaid by the selected token indices from the 2M \ac{vqvae} codebook. 
We highlight the token indices under investigation in red and use blue for the others. 
The input image size is $448 \times 448$ and the patch size is $16 \times 16$, so each image has $28 \times 28$ tokens chosen from the 16384 entries in the codebook. 
First, we highlight the tokens representing the blank white background. 
Comparing the red and blue indices, we observe that the red region has covered most of the blank space inside the table, and the blue region has formulated a convex hull tightly surrounding either the texts or separation lines. 
Next, we highlight the tokens representing the separation lines. 
Comparing the red and blue regions, we observe that the red regions align with all the separations within the table, \eg, line separations and color shading separations. 
Taking a closer look at the indices, we find the codebook has learned to represent an abstract concept by assigning multiple tokens, where each reflects a different scenario.
For example, token ``111'' represents the separation above the gray color shading and token ``964'' represents the one below, and token ``15282'' represents the separation above a bold horizontal line and ``10807'' represents the one below. 
A certain type of separation also has multiple choices depending on the portion of the line within the patch, \eg, token ``14181'', ``8714'', and ``10807'' all representing the ``below a bold horizontal line'', but the differences lie in the line thickness and amount of black inside the image patch. 
Such a fine-grained categorization also explains why the codebook needs so many tokens to represent the implicit conventions created by humans in the table.

\section{Discussions on TSR Dataset Annotations}
\label{appx: annotation}

\begin{figure}
\centering
\begin{subfigure}[b]{0.45\textwidth}
    \centering
    \includegraphics[width=\textwidth]{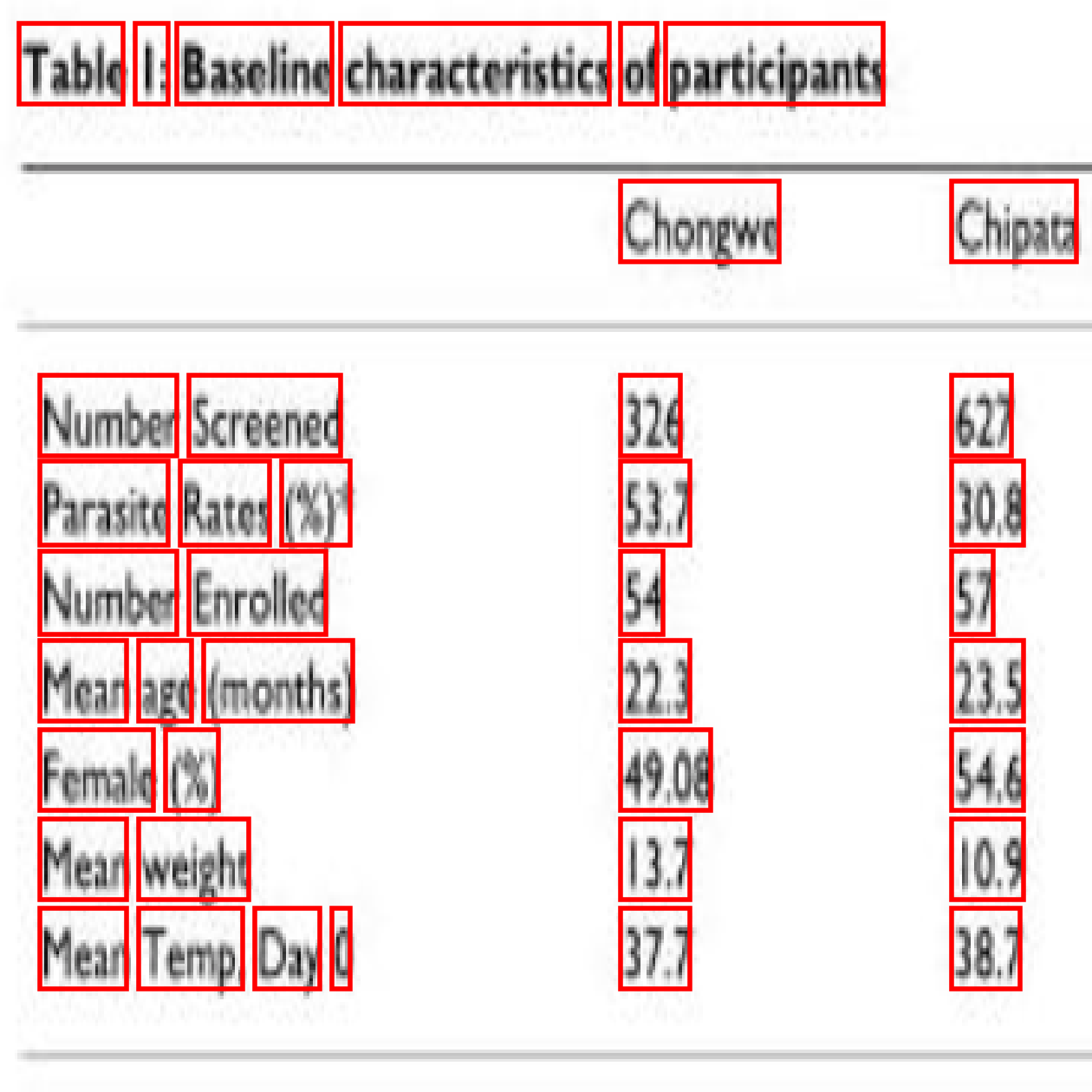}
    \caption{Word-wise \ac{bbox} from PubTables-1M}
    \label{fig:anno-c}
\end{subfigure}
\hfill
\begin{subfigure}[b]{0.45\textwidth}
    \centering
    \includegraphics[width=\textwidth]{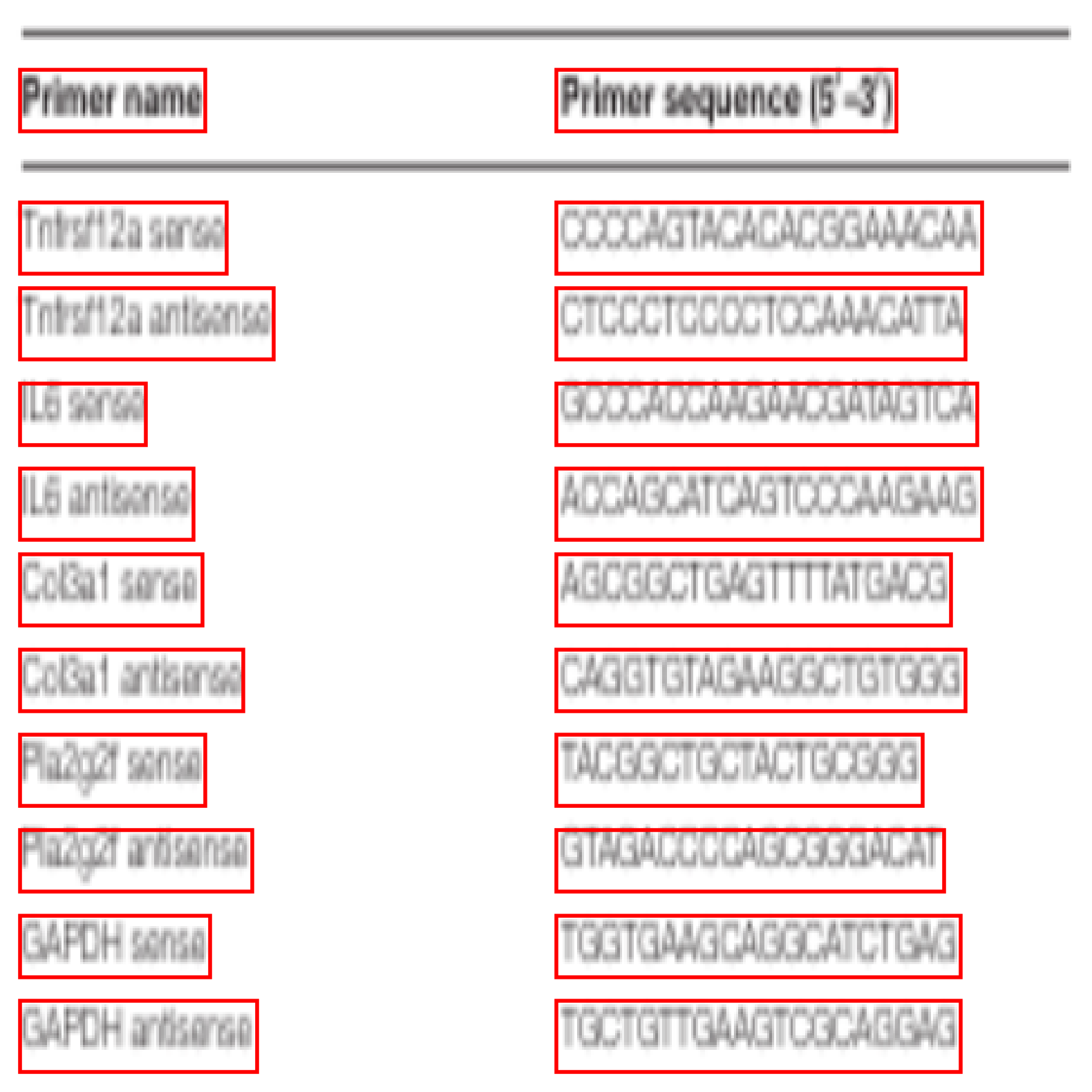}
    \caption{Cell-wise \ac{bbox} from PubTabNet}
    \label{fig:anno-d}
\end{subfigure}
\caption{
Word-wise vs. cell-wise \ac{bbox} annotations: (a) is the word-wise annotation from table ``PMC1574335 table 0'' in PubTables-1M and (b) is the cell-wise annotation from table ``PMC5897438 004 00'' in PubTabNet. 
In cell-wise annotation, each cell has a unique \ac{bbox} that can be matched to the non-empty cells in table structure HTML tags. 
However, in word-wise annotation, a \ac{bbox} is matched to a single word, which cannot be easily combined with the table structure as in cell-wise annotation, which limits its general applicability. 
}
\label{fig:appx-anno-issue}
\end{figure}

\textbf{Word-wise vs. cell-wise \ac{bbox} annotations.} Fig.~\ref{fig:anno-c} is the word-wise annotation from table ``PMC1574335 table 0'' in PubTables-1M and Fig.~\ref{fig:anno-d} is the cell-wise annotation from table ``PMC5897438 004 00'' in PubTabNet. 
In cell-wise annotation, each cell has a unique \ac{bbox} that can be matched to the non-empty cells in table structure HTML tags. 
However, in word-wise annotation, a \ac{bbox} is matched to a single word, which cannot be easily combined with the table structure as in cell-wise annotation, which limits its general applicability.

\subsection{Discussions on TSR Dataset Annotations}
\label{sec: annotation}

\begin{figure}
\centering
\begin{subfigure}[b]{0.45\textwidth}
    \centering
    \includegraphics[width=\textwidth]{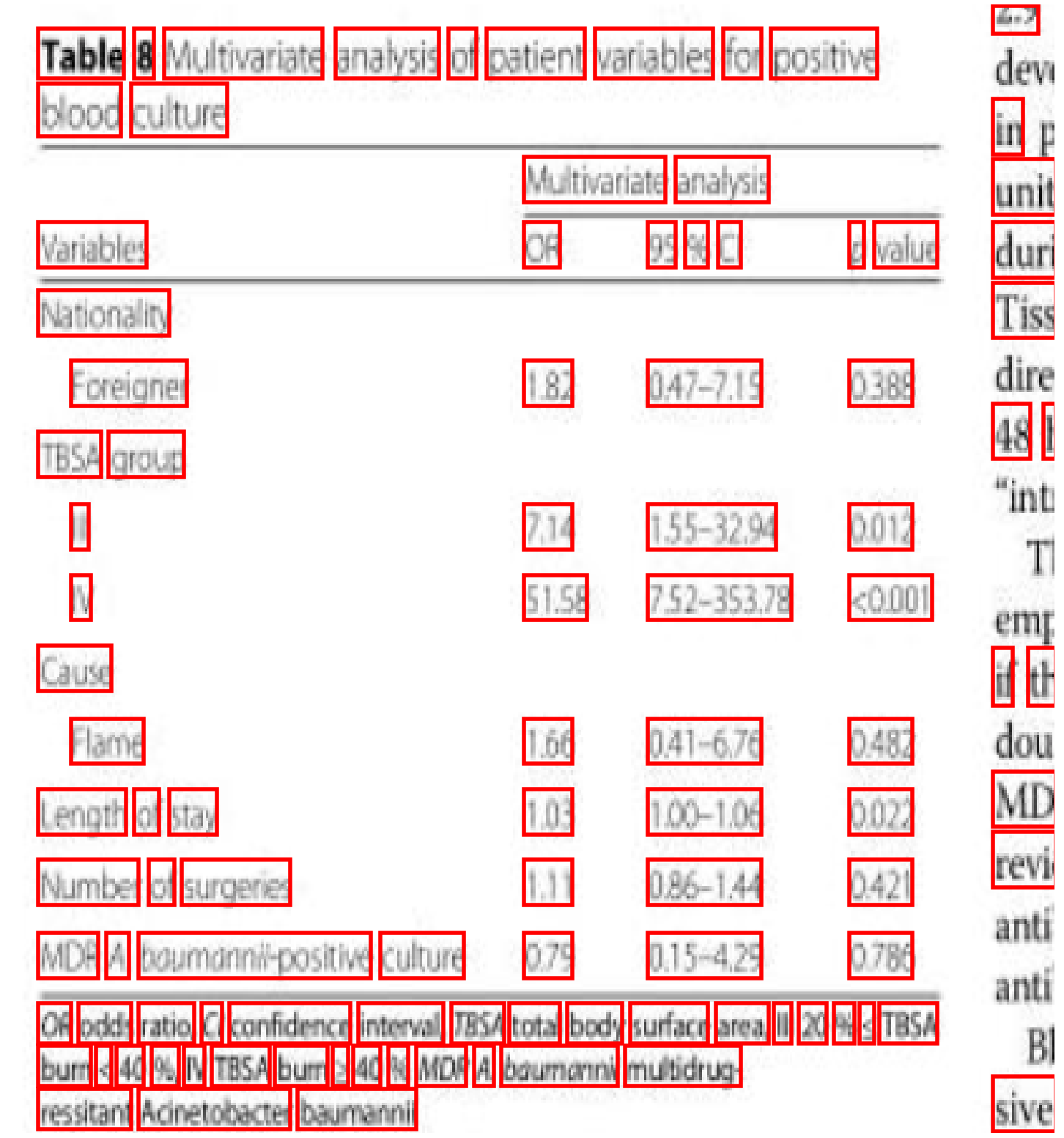}
    \caption{Unrelated text around table}
    \label{fig:anno-a}
\end{subfigure}
\hfill
\begin{subfigure}[b]{0.45\textwidth}
    \centering
    \includegraphics[width=\textwidth]{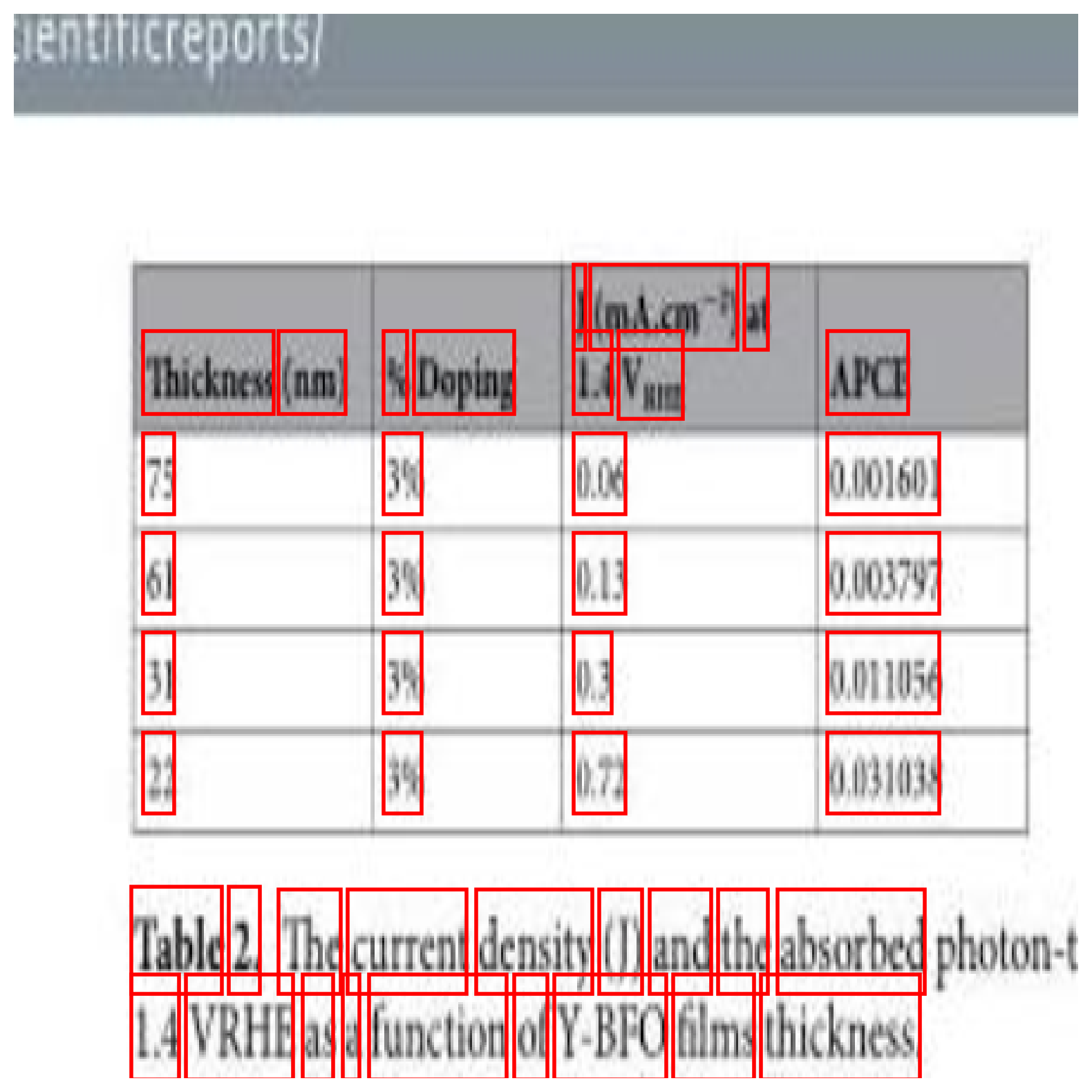}
    \caption{Overlapping \ac{bbox}}
    \label{fig:anno-b}
\end{subfigure}
\caption{Our \method{} has successfully identified inconsistent table cell \ac{bbox} annotations in PubTables-1M, which can guide future data annotation pipelines and qualities.}
\label{fig:anno-issue}
\end{figure}

Our goal is to push forward the development of the document and table understanding, thus we hope to see high-quality and consistent data annotations.
Because of the unified language modeling framework formulation, we discover previously unacknowledged inconsistency in one of the largest \ac{tsr} datasets, PubTables-1M. 

Here, we have identified three types of inconsistent table cell \ac{bbox} annotations:
(1) Fig.~\ref{fig:anno-a} (``PMC4964067 table 7'') shows \ac{bbox} annotations for unrelated text around table;
(2) Fig.~\ref{fig:anno-b} (``PMC6202420 table 1'') shows overlapping \ac{bbox} annotations;
(3) Bbox annotations exceed the boundary of image size, \eg, \texttt{[-4.6, 278.6, 19.5, 292.4]} from ``PMC4802837 table 1''. 
With a rapid filter comparing the image size with the \ac{bbox}, we find that a surprising 53.10\% (402914 over 758849) of the table annotations in the training set have at least one \ac{bbox} going beyond the image size. 
We hope that these inconsistencies discovered by \method{} can help guide future data annotation pipelines and qualities. 








\section{SOTA results}
\label{appx: sota}

\begin{table}[!htbp]
\small
\centering
\caption{Comparisons on IC19B2M}
\begin{tabular}{lrr} 
\toprule
Model & IoU 0.6 & WAvg. F1 \\
\midrule
NLPR-PAL \cite{gao2019icdar} & 30.50 & 20.60 \\
CascadeTabNet \cite{prasad2020cascadetabnet} & 35.40 & 23.20 \\
GTE \cite{zheng2021global} & 38.50 & 24.80 \\
UniTable Base (Ours) & 54.97 & 40.15 \\
UniTable Large (Ours) & \textbf{58.10} & \textbf{42.62} \\
\bottomrule
\end{tabular}
\label{tab:benchmark-1}
\end{table} 
\begin{table}[!htbp]
\small
\centering
\caption{Comparisons on PubTabNet}
\begin{tabular}{lrrrr} 
\toprule
Model & S-TEDS & TEDS & $\text{AP}_{50}$ & $\text{AP}_{75}$ \\
\midrule
EDD \cite{zhong2020image} & 89.90 & 88.30 & 79.20 & - \\
GTE \cite{zheng2021global} & 93.01 & - & - & - \\
TableFormer \cite{nassar2022tableformer} & 96.75 & 93.60 & 82.10 & - \\
TableMaster \cite{ye2021pingan} & 96.04 & 96.16 & - & - \\
OTSL \cite{lysak2023optimized} & 95.50 & - & - & 88.00 \\
VAST \cite{huang2023improving} & 97.23 & 96.31 & 94.80 & - \\
UniTable Base (Ours) & 95.63 & 94.78 & 97.94 & 87.27 \\
UniTable Large (Ours) & \textbf{97.89} & \textbf{96.50} & \textbf{98.43} & \textbf{92.44} \\
\bottomrule
\end{tabular}
\label{tab:benchmark-2}
\end{table}
\begin{table}[!htbp]
\small
\centering
\caption{Comparisons on SynthTabNet}
\begin{tabular}{lrrr} 
\toprule
Model & S-TEDS & $\text{AP}_{50}$ & $\text{AP}_{75}$ \\
\midrule
TableFormer \cite{nassar2022tableformer} & 96.70 & 87.70 & - \\
DRCC \cite{shen2023divide} & 98.70 & - & - \\
UniTable Base (Ours) & 98.97 & 98.99 & 98.79 \\
UniTable Large (Ours) & \textbf{99.39} & \textbf{99.00} & \textbf{98.87} \\
\bottomrule
\end{tabular}
\label{tab:benchmark-3}
\end{table}
\begin{table}[!htbp]
\small
\centering
\caption{Comparisons on FinTabNet}
\begin{tabular}{lr} 
\toprule
Model & S-TEDS \\
\midrule
GTE \cite{zheng2021global} & 91.02 \\
EDD \cite{zhong2020image} & 90.60 \\
OTSL \cite{lysak2023optimized} & 95.90 \\
TableFormer \cite{nassar2022tableformer} & 96.80 \\
VAST \cite{huang2023improving} & 98.63 \\
UniTable Base (Ours) & 97.19 \\
UniTable Large (Ours) & \textbf{98.89} \\
\bottomrule
\end{tabular}
\label{tab:benchmark-4}
\end{table}
\begin{table}[!htbp]
\small
\centering
\caption{Comparisons on PubTables-1M}
\begin{tabular}{lrr}
\toprule
Model & $\text{AP}_{50}$ & $\text{AP}_{75}$ \\
\midrule
Faster R-CNN \cite{smock2022pubtables} & 81.50 & 78.50 \\
OTSL \cite{lysak2023optimized} & - & 89.60 \\
DETR \cite{smock2022pubtables} & \textbf{97.10} & \textbf{94.80} \\
UniTable Base (Ours) & 94.48 & 88.64 \\
UniTable Large (Ours) & 95.68 & 93.28 \\
\bottomrule
\end{tabular}
\label{tab:benchmark-5}
\end{table}

Table~\ref{tab:benchmark-1}, \ref{tab:benchmark-2}, \ref{tab:benchmark-3}, \ref{tab:benchmark-4}, and \ref{tab:benchmark-5} show detailed comparisons across five of the largest table benchmarks from our paper. In summary, \method{} outperforms prior methods and achieves \ac{sota} on four out of the five largest table datasets. (``-'' means not reported by the compared methods.)

\begin{table}
\centering
\caption{Effectiveness and scalability of \ac{ssp} on all four subsets of SynthTabNet. 
Without \ac{ssp}, the model performance suffers, and increasing the model complexity from base to large barely improves the performance.
Here we present both S-\ac{teds} of the table structure prediction and \ac{map} of the cell \ac{bbox} detection
}
\begin{tabular}
{lcccccccc}

\toprule
    & \multicolumn{2}{c}{Finance} & \multicolumn{2}{c}{PubTabNet} & \multicolumn{2}{c}{Marketing} & \multicolumn{2}{c}{Sparse} \\
    & Base & Large & Base & Large & Base & Large & Base & Large \\
\cmidrule(r){2-3}\cmidrule(r){4-5}\cmidrule(r){6-7}\cmidrule(r){8-9}
\multicolumn{5}{l}{\textit{Table structure - S-\ac{teds}}} \\
    No \acs{ssp} & 88.95 & 90.75 & 89.10 & 91.67 & 68.05 & 70.60 & 85.50 & 87.72 \\
    \acs{ssp} 1M & 98.73 & 99.56 & 99.02 & 99.55 & 95.14 & 99.05 & 97.20 & 99.29 \\
    \acs{ssp} 2M & 99.41 & \textbf{99.58} & 99.44 & \textbf{99.56} & 98.35 & \textbf{99.08} & 98.69 & \textbf{99.34} \\
\midrule
\multicolumn{5}{l}{\textit{Table cell \ac{bbox} - COCO \ac{map}}} \\
    No \acs{ssp} & 83.30 & 85.46 & 87.67 & 88.88 & 72.00 & 75.21 & 88.84 & 90.01 \\
    \acs{ssp} 1M & 96.13 & 97.07 & 95.93 & 96.95 & 94.63 & 95.91 & 97.07 & 97.84 \\
    \acs{ssp} 2M & 96.54 & \textbf{97.37} & 96.50 & \textbf{97.44} & 95.21 & \textbf{96.25} & 97.46 & \textbf{97.96} \\
\bottomrule
\end{tabular}
\label{tab:appx-ablation-ssp}
\end{table}

\begin{table}[!htbp]
\centering
\caption{\method{}’s unified training objective applies to both linear projection Transformer and hybrid CNN-Transformer architectures conventionally used in \ac{tsr}. 
Results on all four subsets of the SynthTabNet for table structure prediction evaluated with S-\ac{teds} and cell \ac{bbox} detection with \ac{map}.}
\begin{tabular}{lccc@{\hspace{8mm}}c}
\toprule
    Model & Finance & PubTabNet & Marketing & Sparse \\
\midrule
    \multicolumn{4}{l}{\textit{Table structure - S-\ac{teds}}} \\
    Base  & 98.63 & 98.80 & 97.16 & 95.30 \\
    Large & 99.44 & 99.44 & 98.71 & 98.64 \\
\midrule
    \multicolumn{4}{l}{Table cell \ac{bbox} - COCO \ac{map}} \\
    Base  & 94.61 & 94.39 & 91.42 & 95.38 \\
    Large & 95.90 & 96.50 & 94.96 & 97.55 \\
\bottomrule
\end{tabular}
\label{tab:appx-ablation-unify}
\end{table}

\end{document}